\documentclass[journal, twoside]{./IEEEtran} 

\usepackage[table]{xcolor}
\definecolor{Light1}{rgb}{0.98, 0.95, 0.90}
\definecolor{Light2}{rgb}{0.98, 0.98, 0.93}
\definecolor{Light3}{rgb}{0.98, 0.98, 1}
\definecolor{Light4}{rgb}{0.93, 0.98, 0.98}

\usepackage{graphics} 
\usepackage{epsfig} 
\usepackage{times} 
\usepackage{mathtools}

\usepackage{booktabs}

\usepackage{listings}

\definecolor{pyblue}{RGB}{0,102,204}      
\definecolor{pygreen}{RGB}{0,128,0}       
\definecolor{pyred}{RGB}{187,0,0}         
\definecolor{pypurple}{RGB}{128,0,128}    
\definecolor{pyorange}{RGB}{255,102,0}    
\definecolor{pygray}{RGB}{128,128,128}    
\definecolor{pybackground}{RGB}{248,248,248} 

\usepackage{makecell}

\usepackage{subcaption}

\usepackage{tikz}
\usepackage{ifthen}
\usetikzlibrary{decorations.pathreplacing, calc, backgrounds}
\usetikzlibrary{shapes, shadows.blur, arrows.meta, positioning, arrows, fit}
\usetikzlibrary{automata}
\usepackage{pdfrender}
\usepackage{tikzsymbols}
\usepackage{pgfplots}
\usepackage{pgfplotstable}
\pgfplotsset{compat=1.11}

\usetikzlibrary{arrows.meta, positioning, fit, shapes.symbols, calc}

\makeatletter
\newcommand{\gettikzxy}[3]{%
  \tikz@scan@one@point\pgfutil@firstofone#1\relax
  \edef#2{\the\pgf@x}%
  \edef#3{\the\pgf@y}%
}
\makeatother

\usepackage{multirow}
\usepackage{makecell}

\tikzstyle{generalStyle}=[blur shadow={shadow blur steps=7}]

\newcommand{\drawIndividualFrustum}[7]{
	\set{\xTempFilledBlock}{#1};
	\set{\yTempFilledBlock}{#2};
	\set{\zTempFilledBlock}{-#3};
	
	\set{\leftHeightTempFilledBlock}{#4};
	\set{\rightHeightTempFilledBlock}{#5};
	\set{\widthTempFilledBlock}{#6};

	\pgfmathparse{\leftHeightTempFilledBlock>\rightHeightTempFilledBlock};
	\set{\isLeftBigger}{\pgfmathresult};
	\ifthenelse{\isLeftBigger=1}{
		\set{\leftAdd}{0};
		\set{\rightAdd}{(\leftHeightTempFilledBlock - \rightHeightTempFilledBlock) * 0.5};
	}{
		\set{\leftAdd}{(\rightHeightTempFilledBlock - \leftHeightTempFilledBlock) * 0.5};
		\set{\rightAdd}{0};
	};
	
	\draw [#7] (\xTempFilledBlock, \yTempFilledBlock+\leftAdd, \zTempFilledBlock) -- 
	(\xTempFilledBlock+\widthTempFilledBlock, \yTempFilledBlock + \rightAdd, \zTempFilledBlock) -- 
	(\xTempFilledBlock+\widthTempFilledBlock, \yTempFilledBlock+\rightHeightTempFilledBlock+\rightAdd, \zTempFilledBlock) -- 
	(\xTempFilledBlock, \yTempFilledBlock+\leftHeightTempFilledBlock+\leftAdd, \zTempFilledBlock) -- cycle;	
}

\newcommand{\drawLegendEntry}[6]{
	\pgfmathsetmacro{\posX}{#1};
	\pgfmathsetmacro{\posY}{#2};
	\pgfmathsetmacro{\width}{#3};
	\pgfmathsetmacro{\useit}{#6};
	\pgfmathsetmacro{\posXX}{\posX + \width};
	\node at (\posX,\posY) (start) {};
	\node[anchor=west] at (\posXX,\posY) (end) {#4}; 
	
	\node at (\posX+0.2,\posY-0.17) (startrect) {};
	\node at (\posXX,\posY+0.17) (endrect)  {};
	\begin{scope}
		\ifthenelse{\useit = 1}{
			\draw[#5, line width=3.5mm, draw opacity=0.25] (start) -- (end);
			
			\draw[#5, line width=0.7mm] (start) -- (end);
		}{
			\draw[#5, line width=0.8mm] (start) -- (end);
		};
	\end{scope}
	
}

\definecolor{llmblue}{RGB}{100,150,220}
\definecolor{robotgreen}{RGB}{120,180,120}
\definecolor{frontendgray}{RGB}{200,200,200}
\definecolor{systemgray}{RGB}{240,240,240}
\definecolor{labelred}{RGB}{200,50,50}
\definecolor{arrowblue}{RGB}{70,130,200}
\definecolor{arrowgreen}{RGB}{100,160,100}
\definecolor{arroworange}{RGB}{230,140,50}

\tikzset{
    llmbox/.style={
        rectangle, draw=black, rounded corners=8pt, line width=1.5pt,
        minimum width=3.2cm, minimum height=1.8cm,
        align=center, font=\small\bfseries, fill=llmblue!30
    },
    robotbox/.style={
        rectangle, draw=black, rounded corners=8pt, line width=1.5pt,
        minimum width=3.2cm, minimum height=1.8cm,
        align=center, font=\small\bfseries, fill=robotgreen!30
    },
    frontendbox/.style={
        rectangle, draw=black, rounded corners=8pt, line width=1.2pt,
        minimum width=3cm, minimum height=1.4cm,
        align=center, font=\small, fill=frontendgray!50
    },
    systembox/.style={
        rectangle, draw=black, rounded corners=12pt, line width=2pt,
        fill=systemgray!30, inner sep=15pt
    },
    user/.style={
        circle, draw=black, line width=1.5pt,
        minimum size=1.2cm, fill=white
    },
    robot/.style={
        rectangle, draw=black, line width=1.5pt, rounded corners=3pt,
        minimum width=1.5cm, minimum height=1cm, fill=black
    },
    arrow/.style={-{Latex[width=3mm,length=2mm]}, line width=1.5pt, color=black},
    bluearrow/.style={-{Latex[width=3mm,length=2mm]}, line width=1.5pt, color=arrowblue},
    greenarrow/.style={-{Latex[width=3mm,length=2mm]}, line width=1.5pt, color=arrowgreen},
    orangearrow/.style={-{Latex[width=3mm,length=2mm]}, line width=1.5pt, color=arroworange},
    labeltext/.style={font=\small\bfseries, color=labelred}
}

\tikzstyle{imgStyle}=[blur shadow={shadow blur steps=7, shadow xshift=1pt, shadow yshift=-1pt}]
\tikzset{
	boxFrame/.style={ minimum width=1.5 cm,%
		minimum height=1.5 cm,%
		align=center}
}

\usepackage{xargs} 
\usepackage{caption}
\usepackage{algorithm} 
\usepackage[noend]{algpseudocode}
\usepackage{bm} 
\usepackage{mathbbol}

\usepackage{varwidth}

\newcommand{\set}[2]{\pgfmathsetmacro{#1}{#2}}

\makeatletter
\DeclareRobustCommand\onedot{\futurelet\@let@token\@onedot}
\def\@onedot{\ifx\@let@token.\else.\null\fi\xspace}

\def\eg{\emph{e.g}\onedot}

\def\etal{\emph{et al}\onedot}
\makeatother

\usepackage[breaklinks,colorlinks]{hyperref}

\usepackage[capitalize]{cleveref}
\Crefname{section}{Sec.}{Secs.}
\Crefname{section}{Section}{Sections}
\Crefname{table}{Table}{Tables}
\Crefname{table}{Tab.}{Tabs.}

\usepackage{threeparttable}

\algnewcommand{\LeftComment}[1]{\Statex \(\triangleright\) #1}

\usepackage{cite}
\usepackage{amsmath,amssymb,amsfonts}
\def\BibTeX{{\rm B\kern-.05em{\sc i\kern-.025em b}\kern-.08em
    T\kern-.1667em\lower.7ex\hbox{E}\kern-.125emX}}

\Crefname{equation}{Eq.}{Eqs.} 
\Crefname{figure}{Fig.}{Figs.}

\usepackage{etoolbox}
\makeatletter
\renewcommand\subsubsection{\@startsection{subsubsection}{3}{\z@}%
                                     {-1.2ex \@plus -1ex \@minus -.2ex}%
                                     {-1em}%
                                     {\normalfont\bfseries\color{pyorange}}}
\makeatother

\graphicspath{{}{images-src/}} 

\definecolor{dlrprim1}{HTML}{000000} 
\definecolor{dlrprim2}{HTML}{666666} 
\definecolor{dlrprim3}{HTML}{b9cad2}
\definecolor{dlrprim4}{HTML}{ffffff} 

\definecolor{dlrblue1}{HTML}{00658b} 
\definecolor{dlrblue2}{HTML}{3b98cb}
\definecolor{dlrblue3}{HTML}{6cb9dc}
\definecolor{dlrblue4}{HTML}{a7d3ec}
\definecolor{dlrblue5}{HTML}{d1e8fa}

\definecolor{dlryellow1}{HTML}{d2ae3d}  
\definecolor{dlryellow2}{HTML}{f2cd51} 
\definecolor{dlryellow3}{HTML}{f8de53}
\definecolor{dlryellow3}{HTML}{fcea7a}
\definecolor{dlryellow3}{HTML}{fff8be}

\definecolor{dlrgreen1}{HTML}{82a043} 
\definecolor{dlrgreen2}{HTML}{a6bf51}
\definecolor{dlrgreen3}{HTML}{cad55c}
\definecolor{dlrgreen4}{HTML}{d9df78}
\definecolor{dlrgreen5}{HTML}{e6eaaf}

\definecolor{dlrgray1}{HTML}{666666} 
\definecolor{dlrgray2}{HTML}{868585}
\definecolor{dlrgray3}{HTML}{b1b1b1}
\definecolor{dlrgray4}{HTML}{cfcfcf}
\definecolor{dlrgray5}{HTML}{ebebeb}

\tikzset{
	cross/.pic = {
		\draw[rotate = 45, line width=0.75pt] (-#1,0) -- (#1,0);
		\draw[rotate = 45, line width=0.75pt] (0,-#1) -- (0, #1);
	}
}

\usepackage{orcidlink}

\newcommand{\rmcaffiliation}{German Aerospace Center (DLR),
	Institute of Robotics and Mechatronics (RMC),
	M\"unchener Str. 20, 82234 We\ss ling, Germany. }

\IEEEoverridecommandlockouts                             

\newcommand{\bk}{\bm{k}}
\newcommand{\bK}{\bm{K}}

\newcommand{\mean}{\bm{\mu}}
\newcommand{\covariance}{\bm{\Sigma}}

\newcommand{\inputVariable}{\bm{s}}
\newcommand{\outputVariable}{\bm{\xi}}

\newcommand{\inputDimension}{\mathcal{I}}
\newcommand{\outputDimension}{\mathcal{O}}

\newcommand{\amountOfDatapoints}{M}
\newcommand{\datapointIndex}{m}

\newcommand{\trajectoryLength}{H}
\newcommand{\trajectoryIndex}{h}
\newcommand{\timeInterval}{\delta_t}

\newcommand{\amountOfKMP}{N}
\newcommand{\kmpIndex}{n}

\newcommand{\kernelMatrix}{\bK}
\newcommand{\kernelVector}{\bk^*}
\newcommand{\kernelFunction}{\bk}
\newcommand{\autoKernel}{\kernelFunction^{**}}
\newcommand{\scalarKernel}{k}
\newcommand{\meanVector}{\mean}
\newcommand{\covarianceMatrix}{\covariance}

\newcommand{\viapointMean}{\bar{\mean}}
\newcommand{\viapointCovariance}{\bar{\covariance}}

\newcommand{\viapointInput}{\bar{\inputVariable}}

\newcommand{\kmp}{\bm{D}}
\newcommand{\kmpInput}{\inputVariable_{\kmpIndex}}

\newcommand{\kmpMean}{\mean_{\kmpIndex}}

\newcommand{\kmpCovariance}{\covariance_{\kmpIndex}}

\newcommand{\orcidMarkus}{0000-0001-8229-9410}
\newcommand{\orcidAlin}{0000-0001-5343-9074}
\newcommand{\orcidFreek}{0000-0001-9555-9517}
\newcommand{\orcidJoao}{0000-0003-1428-8933}
\newcommand{\orcidSamuel}{0000-0002-7923-8307}
\newcommand{\orcidThomas}{0000-0002-1074-9504}
\newcommand{\addorcidMarkus}{\orcidlink{\orcidMarkus}}
\newcommand{\addorcidAlin}{\orcidlink{\orcidAlin}}
\newcommand{\addorcidFreek}{\orcidlink{\orcidFreek}}
\newcommand{\addorcidJoao}{\orcidlink{\orcidJoao}}
\newcommand{\addorcidSamuel}{\orcidlink{\orcidSamuel}}
\newcommand{\addorcidThomas}{\orcidlink{\orcidThomas}}

\title{
	IROSA: Interactive Robot Skill Adaptation using Natural Language
}

\author{
	Markus Knauer\addorcidMarkus$^{1, 2}$~(\IEEEmembership{IEEE Student Member}), Samuel Bustamante\addorcidSamuel$^{1, 2}$, Thomas Eiband\addorcidThomas$^{1}$, \\
	Alin Albu-Sch\"affer\addorcidAlin$^{1, 2}$~(\IEEEmembership{IEEE Fellow}), 
	Freek Stulp\addorcidFreek$^{1}$ and Jo\~ao Silv\'erio\addorcidJoao$^{1}$~(\IEEEmembership{IEEE Member}) 
	\thanks{Manuscript received: November, 14, 2025; Revised February, 2, 2026; Accepted February, 27, 2026. This paper was recommended for publication by Editor Jens Kober upon evaluation of the Associate Editor and Reviewers' comments. This work was partially funded by the DLR project "ASPIRO"; the European Union's Horizon Research and Innovation Program under Grant 101136067 (INVERSE); the Federal Ministry for Economic Affairs and Climate Protection with DARP funds based on a decision by the German Bundestag and by the European Union - NextGenerationEU; and partially supported by the German Federal Ministry of Research, Technology and Space (BMFTR) under the Robotics Institute Germany (RIG).}
	\thanks{$^{1}$ All authors are with the \rmcaffiliation \texttt{\{first\}.\{last\}@dlr.de}}
	\thanks{$^{2}$ Markus Knauer, Alin Albu-Sch\"affer and Samuel Bustamante are also with the School of Computation, Information and Technology (CIT), Technical University of Munich (TUM), Arcisstr. 21, 80333 Munich, Germany. \texttt{m.knauer@tum.de}}
	\thanks{Digital Object Identifier (DOI): \href{https://doi.org/10.1109/LRA.2026.3671560}{10.1109/LRA.2026.3671560}}
	\thanks{Code available at: \url{https://github.com/DLR-RM/IROSA}}
	
}

\begin{document}

	\maketitle

	\begin{abstract}
		Foundation models have demonstrated impressive capabilities across diverse domains, while imitation learning provides principled methods for robot skill adaptation from limited data.
Combining these approaches holds significant promise for direct application to robotics, yet this combination has received limited attention, particularly for industrial deployment.
We present a novel framework that enables open-vocabulary skill adaptation through a tool-based architecture maintaining a protective abstraction layer between the language model and robot hardware.
Our approach leverages pre-trained LLMs to select and parameterize specific tools for adapting robot skills without requiring fine-tuning or direct model-to-robot interaction.
We demonstrate the framework on a 7-DoF torque-controlled robot performing an industrial bearing ring insertion task, showing successful skill adaptation through natural language commands for speed adjustment, trajectory correction, and obstacle avoidance while maintaining safety, transparency, and interpretability.
	\end{abstract}
	
	\begin{IEEEkeywords}
		Foundation Models, Imitation Learning
	\end{IEEEkeywords}
	
	\section{Introduction}
	\IEEEPARstart{R}{obotics} applications increasingly demand flexible, adaptive systems that can be easily reconfigured for varying tasks and environments.
Large language models (LLMs) offer significant potential for making robot skill adaptation more accessible to non-expert users, enabling intuitive control through everyday language rather than specialized programming.
Our aim is to apply this capability to industrial tasks, where interpretable, verifiable, and predictable robot behaviors are essential.
This requires an approach that preserves established control principles while enabling natural language adaptation.

\begin{figure}
	\centering
	\includegraphics[width=1.0\columnwidth]{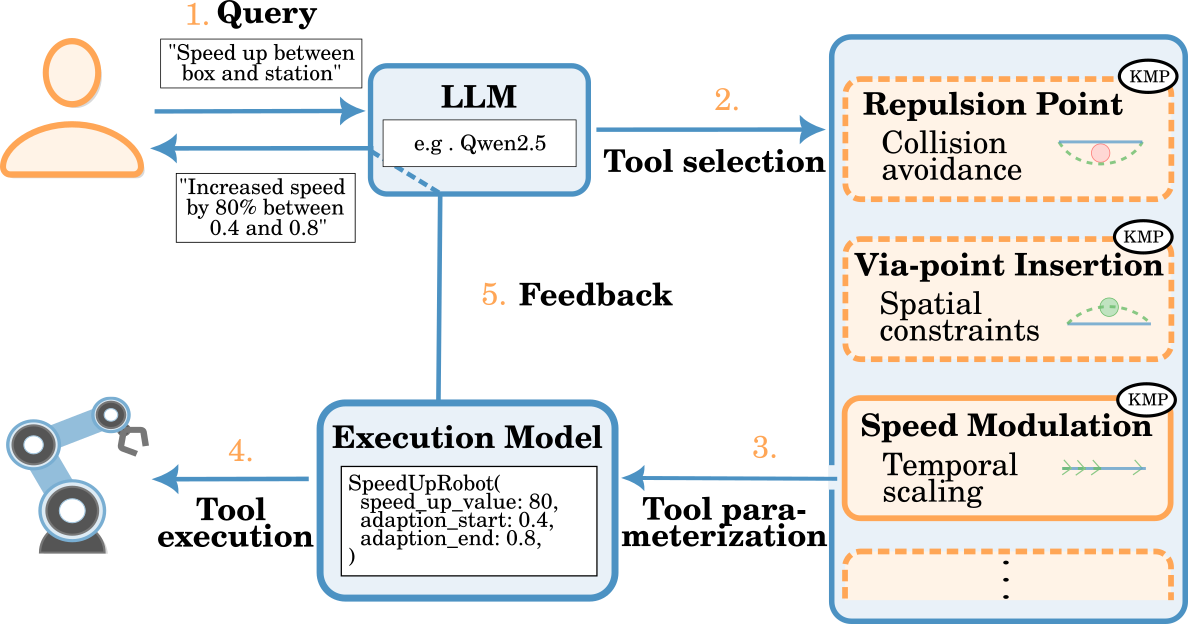}
	\caption{Overview of our approach \textbf{I}nteractive \textbf{RO}bot \textbf{S}kill \textbf{A}daptation using natural language. Showing the interactive selection and parameterization of a tool by a LLM based on a user query leading to a skill adaptation via the used execution model. Some of the tools we are providing are shown. "Respond to User" is a general tool, whereas "Repulsion Point", "Via-Point Insertion" and "Speed Modulation" are specific tools to adapt KMPs.}
	\label{fig:graphical_abstract}
\end{figure}
%
%
%
We address this challenge through a tool-based architecture maintaining strict separation between language understanding and robot control.
Our approach builds on \textit{function calling}~\cite{OpenAI2023, Schick2023, Qin24}, where language models select and parameterize predefined functions based on natural language input rather than directly generating outputs or actions.
We provide LLMs with a well-defined toolbox of modification primitives, where each tool represents a validated, parameterized function that modifies an underlying motion generation model.
The LLM selects and parameterizes tools based solely on their textual descriptions, enabling zero-shot adaptation without retraining or fine-tuning.

We map these adaptation primitives to a trajectory generation framework flexible enough to accommodate diverse modifications while maintaining deterministic, interpretable behavior.
We realize this through Kernelized Movement Primitives (KMPs)~\cite{Huang19}---a non-parametric probabilistic imitation learning framework providing temporal and spatial adaptation capabilities through start-, end-, and via-points (see \Cref{sec:preliminaries}).
Rather than allowing LLMs direct access to robot hardware or trajectory generation, our framework constrains the language model to select and parameterize well-defined, validated robot control functions.

This design contrasts with end-to-end approaches that train models directly on sensorimotor data.
While such approaches (see \Cref{sec:relatedWork}) demonstrate impressive capabilities in some domains, they face challenges in meeting the interpretability, verifiability, and safety compliance requirements of industrial applications.
Our approach leverages pre-trained LLMs combined with established probabilistic motion representations, providing a cost-effective, explainable, and immediately deployable solution.
%
Our key contributions (\Cref{sec:approach}) are:
\begin{enumerate}
    \item A tool-based architecture that enables zero-shot natural language adaptation of robot skills through structured function calling, maintaining strict separation between language understanding and robot control.
    \item Novel extensions to KMPs for natural language-driven speed modulation and obstacle avoidance via repulsion fields, expanding the adaptation capabilities beyond traditional via-point constraints.
    \item Experimental validation (\Cref{sec:evaluation}) on a 7-DoF torque-controlled DLR SARA robot \cite{Iskandar20} performing industrial manipulation tasks, demonstrating reliable and explainable skill adaptation without model fine-tuning or iterative user feedback.
\end{enumerate}

	\section{Related work}
	\label{sec:relatedWork}

\textbf{Probabilistic representations of movement primitives:} Gaussian Mixture Models (GMMs)~\cite{Calinon07} capture uncertainty but struggle with via-point adaptation after learning.
Probabilistic Movement Primitives (ProMPs)~\cite{Paraschos13} enable via-point adaptation but require basis function definition, becoming cumbersome in high-dimensional spaces.
Kernelized Movement Primitives (KMPs)~\cite{Huang19} overcome these limitations through a non-parametric kernel-based approach that trivially accommodates via-point constraints. Task-Parameterized KMPs (TP-KMPs)~\cite{Calinon16} further enable adaptation based on task-relevant coordinate frames.
Recently, Knauer \etal~\cite{Knauer2025} introduced an interactive incremental learning framework combining TP-KMPs with kinesthetic human feedback. This approach enables real-time skill modification through physical human-robot interaction, where users guide the robot's end-effector to demonstrate desired trajectory corrections.
While kinesthetic feedback excels at fine-grained trajectory adjustments, it becomes cumbersome for larger-scale modifications or when specifying abstract behavioral changes such as ``avoid the obstacle'' or ``move faster.''

\textbf{End-to-End Language-Conditioned Control:} Recent approaches explore direct integration of language conditioning into robot control policies. CLIPORT~\cite{shridhar2021} combines CLIP-based semantic understanding with transporter networks for language-conditioned manipulation. KITE~\cite{Sundaresan2023} extends this through keypoint-conditioned policies that enable fine-grained manipulation based on object features. LaTTe~\cite{Bucker2023} follows an \textit{end-to-end} approach, employing a pre-trained transformer to directly map natural language instructions to trajectory modifications. While LaTTe achieves fast inference ($<1$ second), it is limited to instruction patterns seen during training.
Vision-language-action (VLA) models represent the current state-of-the-art. OpenVLA~\cite{Kim2025} trains large-scale policies on diverse robot datasets for generalized manipulation, while RoboPoint~\cite{Yuan2025} predicts spatial affordances based on natural language commands. While these approaches demonstrate impressive capabilities, their direct application to robot control faces significant challenges.
Particularly, they require extensive training data, lack interpretability, and provide limited safety guarantees for industrial settings~\cite{Ravichandar20, Celemin22, Neill_2024}.

\textbf{Modular Language-Robot Interfaces:} Modular architectures decouple natural language understanding from low-level motion generation, enabling independent validation of each component. These state-of-the-art solutions differ in \emph{how} the LLM interfaces with the robot control layer.
OLAF~\cite{Huihan2023} follows a \textit{policy retraining} approach, separating language understanding from visuomotor policy execution. The LLM generates policy update signals rather than direct trajectory modifications, but this requires \emph{retraining} the policy network after each verbal correction.
Sharma et al.~\cite{Sharma2022} decouple language processing from motion planning by learning a mapping from language to cost function modifications. However, this requires \emph{offline training}, and cost modifications are applied before replanning rather than enabling real-time adaptation.
Merlo et al.~\cite{Merlo2024} separate vision-based plan generation from LLM-based replanning, using the LLM for common-sense reasoning about potential failure modes. The LLM operates at the \emph{action/task level} rather than the trajectory level, adjusting symbolic plans rather than continuous waypoints.
Yu et al.~\cite{Yu2023} separate LLM-based reward specification from policy optimization, where the LLM generates reward function parameters that are then optimized via MPC in simulation. While this improves stability, it \emph{requires simulation-based optimization} rather than direct trajectory adaptation.
OVITA~\cite{Maurya2025} follows a \textit{code generation} approach, separating language understanding from trajectory execution through LLM-generated Python code that modifies waypoints. While flexible, this inherits \emph{code generation reliability issues}, requires \emph{cloud-based LLM APIs}, and relies on \emph{iterative user feedback} for validation.
Diffusion models have also been explored for language-based skill adaptation~\cite{Kim2024}. In contrast, KMPs provide analytical uncertainty expressions and require fewer demonstrations to learn. 
%

\textbf{Positioning of Our Approach:} The modular approaches above share common limitations: OLAF and Sharma et al.\ require retraining or offline learning after corrections; Merlo et al.\ operate at the symbolic action level rather than on continuous trajectories; Yu et al.\ require simulation-based optimization loops; and OVITA's code generation introduces reliability issues and cloud dependency.
Our \textit{tool-based} approach addresses these limitations by combining LLMs with Kernelized Movement Primitives, leveraging KMPs' flexibility to adapt learned skills to new situations through via-points and temporal modulation. The LLM is constrained to selecting and parameterizing predefined tools, providing safety guarantees through validated tool functions. This enables direct trajectory-level adaptation without retraining, simulation, or iterative code validation, while supporting local LLM deployment for offline industrial environments.
A summary comparing our approach with related work is shown in \cref{tab:approach_comparison}.
\begin{table}[htbp]
\centering
\caption{Comparison of Language-Conditioned Trajectory Adaptation Approaches. Training-Free: no task-specific (re)training required. Open Vocabulary: handles arbitrary natural language instructions. Zero-shot: no iterative user feedback required to validate model output. Explainable: adaptation steps are traceable and interpretable. Local LLM: operates with locally-deployed language models.}
\label{tab:approach_comparison}
\resizebox{\columnwidth}{!}{%
\begin{tabular}{|l|c|c|c|c|c|c|}
\hline
\textbf{Method} & \textbf{Approach} & \makecell{\textbf{Training-}\\\textbf{Free}} & \makecell{\textbf{Open}\\\textbf{Vocab.}} & \makecell{\textbf{Zero-}\\\textbf{Shot}} & \makecell{\textbf{Explain-}\\\textbf{able}} & \makecell{\textbf{Local}\\\textbf{LLM}} \\
\hline
OLAF \cite{Huihan2023} & Policy retraining & $\times$ & $\sim$ & $\times$ & $\times$ & \checkmark \\
\hline
LaTTe \cite{Bucker2023} & End-to-end & $\times$ & $\times$ & \checkmark & $\times$ & \checkmark \\
\hline
Merlo \cite{Merlo2024} & Action Replan. & \checkmark & \checkmark & $\times$ & \checkmark & $\sim$ \\
\hline
Yu \cite{Yu2023} & Reward Synth. & $\times$ & \checkmark & $\times$ & $\sim$ & $\times$ \\
\hline
OVITA \cite{Maurya2025} & Code Gen. & \checkmark & \checkmark & $\times$ & \checkmark & $\times$ \\
\hline
\textbf{Ours} & Tool-based & \checkmark & \checkmark & \checkmark & \checkmark & \checkmark \\
\hline
\end{tabular}
}
\end{table}

	\section{Preliminaries}
	\label{sec:preliminaries}
	Our approach begins with a set of demonstrations $\mathcal{D} = \{\{\inputVariable_{\trajectoryIndex,\datapointIndex},\outputVariable_{\trajectoryIndex,\datapointIndex}\}_{\trajectoryIndex=1}^\trajectoryLength\}_{\datapointIndex=1}^\amountOfDatapoints$, where:
\begin{itemize}
	\item $\inputVariable\in\mathbb{R}^{\inputDimension}$ represents the input variable (typically time or phase variable)
	\item $\outputVariable \in \mathbb{R}^{\outputDimension}$ represents the output variable (e.g., end-effector position, joint angles)
	\item $\inputDimension$, $\outputDimension$ denote the dimensions of input and output spaces
	\item $\amountOfDatapoints$ is the number of demonstrations
	\item $\trajectoryLength$ is the trajectory length (number of datapoints per demonstration)
\end{itemize}

Following established LfD approaches~\cite{Calinon07, Huang19, Paraschos13}, we extract the relationship between $\inputVariable$ and $\outputVariable$ from demonstrations obtained through kinesthetic teaching.

\textbf{Kernelized Movement Primitives (KMPs)}~\cite{Huang19} provide a probabilistic approach for learning robot skills from limited human demonstrations.
The key advantage of KMPs for our application is their ability to learn meaningful skill representations from small datasets (2--5 demonstrations) while enabling principled trajectory modifications through constraint incorporation.
From the demonstrations $\mathcal{D}$, a KMP encodes the skill as a \textit{reference trajectory distribution} $\kmp = \left\lbrace \kmpInput, \kmpMean, \kmpCovariance \right\rbrace^\amountOfKMP_{\kmpIndex=1}$ comprising $\amountOfKMP$ Gaussians with mean and covariance parameters, computed for inputs $\inputVariable_{\kmpIndex=1,\dots,\amountOfKMP}$ using Gaussian Mixture Models (GMMs).
For a query input $\inputVariable^*$ (typically time), the expectation and covariance of the predicted output $\outputVariable(\inputVariable^*)$ 
are given by \cref{eq:kmp_mean}--\cref{eq:kmp_cov}
\begin{align}
	\mathbb{E}\left[\outputVariable(\inputVariable^*)\right] \! &= \! \kernelVector\left(\kernelMatrix + \lambda_1\covarianceMatrix\right)^{-1}\meanVector,\label{eq:kmp_mean}\\
	\mathrm{cov}\left[\outputVariable(\inputVariable^*)\right] \! &= \!\alpha\left(\autoKernel-\kernelVector\left(\kernelMatrix + \lambda_2\covarianceMatrix\right)^{-1}{\kernelVector}^{\top}\right),\label{eq:kmp_cov}
\end{align}
where $\kernelMatrix = [\hat{\kernelFunction}(\inputVariable_1)^\top,\ldots,\hat{\kernelFunction}(\inputVariable_\amountOfKMP)^\top]$, $\kernelVector= \hat{\kernelFunction}(\inputVariable^*)$, with $\hat{\kernelFunction}(\inputVariable_{\kmpIndex}) = [\kernelFunction(\inputVariable_{\kmpIndex},\inputVariable_1),\ldots,\kernelFunction(\inputVariable_{\kmpIndex},\inputVariable_\amountOfKMP)]$, $\autoKernel = \kernelFunction(\inputVariable^*,\inputVariable^*)$, $\kernelFunction(\inputVariable_{\kmpIndex},\inputVariable_{\kmpIndex'})=\scalarKernel(\inputVariable_{\kmpIndex},\inputVariable_{\kmpIndex'})\bm{I}$, where $\bm{I}$ is an identity matrix, and $\scalarKernel(\inputVariable_{\kmpIndex},\inputVariable_{\kmpIndex'})$ represents a scalar kernel function. The mean and covariance vectors are $\meanVector =  \left[\mean_{\kmpIndex}^\top\right]_{\kmpIndex=1}^\amountOfKMP$ and $\covarianceMatrix = \mathrm{blockdiag}\left(\covariance_{\kmpIndex}\right)_{\kmpIndex=1}^\amountOfKMP$, respectively, while $\lambda_1$, $\lambda_2$, and $\alpha$ are hyperparameters controlling regularization and scaling.
A key property of KMPs is their ability to incorporate new constraints through via-points. From \cref{eq:kmp_mean,eq:kmp_cov}, if the covariance $\kmpCovariance$ is small for a particular $\kmpMean$, the predicted expectation at $\kmpInput$ will be close to $\kmpMean$. This property enables principled trajectory modulation: to ensure that the predicted trajectory passes through a desired point $\viapointMean$ at input $\viapointInput$, one can add the via-point triplet $\{\viapointInput, \viapointMean,\viapointCovariance\}$ to the reference distribution, provided that $\viapointCovariance$ is sufficiently small. This modification forces \cref{eq:kmp_mean} to closely match $\viapointMean$ while reducing the prediction uncertainty in \cref{eq:kmp_cov}.
We leverage KMPs' kernel formulation that trivially accommodates via-point constraints to enable our tool-based framework to seamlessly incorporate natural language-specified trajectory modifications without requiring explicit basis function definitions.

	\section{Approach}
	\label{sec:approach}
	Our approach enables users to adapt robot skills through natural language commands, supporting three primary adaptation types relevant for industrial tasks:
\textbf{Speed Modulation} enables temporal adjustments to trajectory execution (\eg ``move faster before reaching the box,'' ``slow down by 50\%''), addressing varying task requirements such as adapting to different production speeds or ensuring careful approach to delicate objects.
\textbf{Via-point Insertion} provides spatial trajectory modifications (\eg ``move 10cm to the left,'' ``approach from above''), enabling users to correct trajectories based on changed workspace layouts or to avoid interference with other equipment.
\textbf{Repulsion Point Generation} supports collision avoidance through obstacle awareness (\eg ``avoid the blue box,'' ``stay away from the red object''), accommodating dynamic workspace changes where new obstacles appear or existing objects are relocated.

These adaptations are motivated by common industrial scenarios where robots must accommodate variations in object positions, production rates, and workspace configurations without requiring expert reprogramming.

\subsection{Problem Formulation}

Our goal is to adapt a robot skill based on natural language instructions while preserving the demonstrated skill structure and ensuring safe, predictable robot behavior. Given a learned KMP $\kmp$ from demonstrations $\mathcal{D}$ (see \Cref{sec:preliminaries}), the system receives: (i) the \textbf{initial KMP} $\kmp$ encoding the demonstrated skill, (ii) a \textbf{user instruction} $L_{\text{instruct}}$ in natural language, (iii) \textbf{environment observations} $\mathcal{E} = \{\bm{p}_{i}, \bm{d}_{i}, \ell_{i}\}_{i=1}^{N_{\text{obj}}}$ where $N_{\text{obj}}$ is the number of objects, $\bm{p}_{i} \in \mathbb{R}^3$ is the object position, $\bm{d}_{i} \in \mathbb{R}^3$ are object dimensions, and $\ell_{i}$ is the object semantic label, and (iv) an optional \textbf{environment description} $E_d$ in natural language.
In the current implementation, environment observations are provided manually, which is common in structured industrial environments. Integration with automatic perception systems (\eg pose estimation~\cite{Sundermeyer2018}) is straightforward.
The KMP framework supports full 6-DoF manipulation, encoding both position and orientation; our evaluation focuses on position adaptations, with an orientation modification example demonstrated in the supplementary video.
The system modifies the KMP's internal trajectory representation according to $L_{\text{instruct}}$ and $\mathcal{E}$ through the five-step workflow detailed below.

\subsection{Five-Step Workflow}
The system processes commands through the workflow illustrated in \Cref{fig:graphical_abstract}: (1) User Query, (2) Tool Selection, (3) Tool Parameterization, (4) Tool Execution, and (5) Feedback.

\subsubsection{User Query}
The robot executes the learned skill using the KMP $\kmp$. Users observe the execution and provide natural language instructions $L_{\text{instruct}}$ via voice commands or text input when adaptation is desired. These instructions can be simple commands (\eg ``move faster'') or more complex specifications (\eg ``slow down by 50\% before reaching the box''). The system also receives environment observations $\mathcal{E}$ and optional environment description $E_d$ to provide context for instruction interpretation.

\subsubsection{Tool Selection}

The LLM analyzes the user instruction $L_{\text{instruct}}$ and environment observations $\mathcal{E}$ to select appropriate tool(s) $T^* \in \mathcal{T}$ from the available tool set $\mathcal{T}$. This selection leverages the LLM's native function calling capabilities without requiring task-specific training. All available tools are serialized into JSON schemas containing their name, description, and parameter specifications with type annotations. This follows the standard JSON schema-based tool definition format widely adopted across LLM APIs \cite{OpenAI2023, Schick2023, Qin24}. These schemas are sent to the LLM alongside the user's natural language command, and the LLM directly returns the selected tool(s) with instantiated parameters in a single inference step, leveraging the LLM's internal semantic understanding to match user intent with tool descriptions.

\textbf{Tool Descriptions:} Each tool includes a description specifying its function and parameters. For example, the \texttt{SpeedUpRobot} tool is defined as:

\begin{lstlisting}[
    language=Python,
    caption=Example tool definition with natural language description,
    basicstyle=\tiny\ttfamily\color{black},
    backgroundcolor=\color{pybackground},
    keywordstyle=\color{pyblue}\bfseries,
    commentstyle=\color{pygreen}\itshape,
    stringstyle=\color{pyred},
    numberstyle=\tiny\color{pygray},
    identifierstyle=\color{black},
    emphstyle=\color{pypurple}\bfseries,
    emph={Tool, int, float, str, Exception, ToolArgsValidationError},
    emph={[2]self, robot, kmp},
    emphstyle={[2]\color{pyorange}\itshape},
    breaklines=true,
    breakatwhitespace=true,
    columns=flexible,
    keepspaces=true,
    showstringspaces=false,
    tabsize=4,
    frame=single,
    frameround=tttt,
    framesep=5pt,
    rulecolor=\color{pygray!50},
    captionpos=b,
    numbers=left,
    stepnumber=1,
    numbersep=8pt,
    xleftmargin=17pt,
    xrightmargin=5pt
]
SpeedUpRobot(speed_up_value: int, adaption_start: float, 
             adaption_end: float):
    """
    Detailed tool description for the LLM.
    Containing parameter description. E.g.
    :param speed_up_value: percentual speedup of the robot.
    ...
    """
    
    # Type and range validation for all parameters
    if adaption_start >= adaption_end:
        raise ValidationError("adaption_start must be less than
                              adaption_end")
	    ...

	#Tool execution
    try:
        #kmp: KMPWrapper is a global variable
        return kmp.change_predicting_frequency(speed_up_value, 
                                  adaption_start, adaption_end)
    except Exception as e:
        raise ExecutionError("Error speeding up robot: " + str(e))

\end{lstlisting}

These descriptions serve as the interface between human intent and machine execution, allowing the LLM to match user requests with appropriate tools.

\subsubsection{Tool Parameterization}

Once the appropriate tool $T^*$ is selected, the LLM extracts parameters $\theta^*$ for the selected tool based on $L_{\text{instruct}}$, $\mathcal{E}$, and the tool description. For example, if the user command is ``slow down by 50\% before reaching the box'', the LLM parameterizes the \texttt{SpeedUpRobot} tool with: \texttt{speed\_up\_value}$=50$ (indicating slowdown), \texttt{adaption\_start}$=0.0$, and \texttt{adaption\_end} determined based on the box position in $\mathcal{E}$.

\textbf{Trajectory segment determination} follows a two-step process: (1) the LLM identifies the relevant spatial references from the user's request (\eg ``box'' and ``station''); (2) inside the tool function, the system iterates over the predicted trajectory and returns the closest time point for each referenced object (within the hard-coded proximity threshold $d_{\text{prox}}$). When multiple objects are specified (\eg ``between box and station''), these time points define $t_{\text{start}}$ and $t_{\text{end}}$. This constrains the LLM to semantic understanding while delegating numerical computation to deterministic algorithms inside the tool.

\textbf{Parameter validation} provides multiple safety layers to ensure tool execution remains within safe operational bounds. These validation checks are implemented within each tool's execution method: \textit{type validation} verifies that extracted parameters match expected data types (\eg numerical values for speed adjustments), and \textit{range validation} confirms parameters fall within pre-defined safe operational bounds (\eg speed changes within $\pm$200\% of baseline).
When validation fails, the system provides specific feedback to guide user refinement, maintaining transparent operation.

\subsubsection{Tool Execution}

Once the tool $T^*$ and parameters $\theta^*$ are validated, the tool executes and modifies the KMP's internal representation: $\kmp \gets T^*(\kmp, \theta^*)$. The KMP model serves as the foundation for our tool-based interface, providing three primary adaptation mechanisms that leverage the inherent flexibility of KMPs:

\begin{itemize}
	\item \textbf{Speed Modulation:} Adjusts the execution speed of trajectory segments by modifying the time intervals between trajectory points. The tool parameters are $\theta = \{\inputVariable_{\text{start}}, \inputVariable_{\text{end}}, \gamma\}$: trajectory segment start/end (normalized time inputs) and percentage speed change. Given a baseline trajectory with time steps ${\timeInterval}_{\trajectoryIndex}$ between consecutive points at normalized time inputs $\inputVariable_{\trajectoryIndex} \in [0,1]$, we scale the time intervals within a segment $[\inputVariable_{\text{start}}, \inputVariable_{\text{end}}]$ by a factor $f(\gamma)$:
	\begin{align}
	{\timeInterval'}_{\trajectoryIndex} &= \begin{cases}
	{\timeInterval}_{\trajectoryIndex} \cdot f(\gamma) & \text{if } \inputVariable_{\text{start}} \leq \inputVariable_{\trajectoryIndex} < \inputVariable_{\text{end}} \\
	{\timeInterval}_{\trajectoryIndex} & \text{otherwise}
	\end{cases}\\
	f(\gamma) &= \begin{cases}
	\frac{|\gamma| + 100}{100} & \text{if } \gamma \geq 0 \\
	\frac{100}{|\gamma| + 100} & \text{if } \gamma < 0
	\end{cases}
	\end{align}
	Here $\gamma > 0$ indicates slowdown and $\gamma < 0$ indicates speedup. For example, $\gamma = 50$ increases time intervals by $1.5\times$, slowing execution by 50\%, while $\gamma = -50$ decreases intervals to $0.67\times$, speeding up by 50\%. Points outside the specified segment maintain their original timing. This enables commands such as ``move faster before reaching the box'' or ``slow down by 50\%,'' while preserving the spatial characteristics of the learned skill.
	\item \textbf{Via-point Insertion:} Adds via-points $\{\viapointInput, \viapointMean, \viapointCovariance\}$ to steer the trajectory through desired regions (see \Cref{sec:preliminaries}). The parameters $\theta = \{\mathbf{p}_i, s_i, \gamma_\Sigma\}$ specify via-point positions, normalized time inputs, and covariance scaling. The number of via-points is determined by the desired minimum dwell time at the target, a hard-coded system parameter. This supports commands like ``move more to the left'' or ``approach from above''.
	\item \textbf{Repulsion Point Generation:} Introduces repulsive constraints that encourage the robot to avoid specified regions in the workspace. The tool parameters are $\theta = \{\bm{c}, r_{\text{obs}}, \delta_{\text{safe}}\}$: obstacle center position, radius, and safety margin. Given an obstacle with center $\bm{c} \in \mathbb{R}^3$ and radius $r_{\text{obs}}$, we define a signed distance field (SDF)~\cite{Osher1988}
	\begin{equation}
	d(\bm{p}) = \|\bm{p} - \bm{c}\| - r_{\text{obs}},
	\end{equation}
	where $\bm{p} \in \mathbb{R}^3$ is a point in workspace. While we use a spherical SDF for simplicity, the SDF representation enables arbitrary obstacle shapes, allowing exact object geometries to be incorporated when available, as in our experiments (\Cref{subsec:obstacle_avoidance}). The predicted trajectory is checked for collision at discrete inputs $\inputVariable_{\trajectoryIndex}$, where $\mathbb{E}[\bm{\xi}(\inputVariable_{\trajectoryIndex})]$ denotes the expected position from the KMP at input $\inputVariable_{\trajectoryIndex}$ (Eq.~\eqref{eq:kmp_mean}): if $d(\mathbb{E}[\bm{\xi}(\inputVariable_{\trajectoryIndex})]) < \delta_{\text{safe}}$ for safety margin $\delta_{\text{safe}}$, we compute corrected positions $\bm{x}'(\inputVariable_{\trajectoryIndex})$ by pushing points outside the obstacle:
	\begin{equation}
	\bm{x}'(\inputVariable_{\trajectoryIndex}) = \bm{c} + \frac{\mathbb{E}[\bm{\xi}(\inputVariable_{\trajectoryIndex})] - \bm{c}}{\|\mathbb{E}[\bm{\xi}(\inputVariable_{\trajectoryIndex})] - \bm{c}\|} \cdot (r_{\text{obs}} + \delta_{\text{safe}}).
	\end{equation}
	Via-points are placed at regular intervals along the affected trajectory segment: for all indices where $\|\mathbb{E}[\bm{\xi}(\inputVariable_{\trajectoryIndex})] - \bm{x}'(\inputVariable_{\trajectoryIndex})\| > \varepsilon_{\text{thresh}}$ with threshold $\varepsilon_{\text{thresh}}$, we add via-points $\{\viapointInput_{\trajectoryIndex} = \inputVariable_{\trajectoryIndex}, \viapointMean_{\trajectoryIndex} = [\bm{x}'(\inputVariable_{\trajectoryIndex}), \bm{q}(\inputVariable_{\trajectoryIndex})], \viapointCovariance_{\trajectoryIndex}\}$ to the KMP. The number of via-points depends on the collision segment length; we select representative points to ensure smooth trajectory deformation while keeping modifications localized. This leverages the via-point property of KMPs (\Cref{sec:preliminaries}) to generate collision-free trajectories, enabling commands like ``avoid the blue box''.

\end{itemize}

These adaptation mechanisms leverage the flexibility of KMPs to incorporate new constraints while maintaining the learned skill structure, ensuring that adaptations remain consistent with the original demonstrations. Due to space constraints, we only present the key functionalities, but other functions were also implemented, including hyperparameter setting via natural language.

\subsubsection{Feedback and Iterative Refinement}

After the modified KMP is executed by the robot, the system provides feedback to the user and awaits further instructions. This enables iterative refinement where users can observe the adapted behavior and provide additional commands to fine-tune the skill. The robot executes the modified skill using the updated $\kmp$ and the cycle returns to Step 1, allowing continuous adaptation based on user feedback.

Key advantages of this architecture include: \textbf{(1) Safety through Abstraction:} the LLM never directly controls robot trajectories but interfaces only with well-defined, deterministic tools that implement safe modifications to the underlying model, helping mitigate potential issues from LLM hallucinations while ensuring predictable robot behavior. \textbf{(2) Interpretability and Transparency:} unlike end-to-end learned policies where decision-making lacks transparency, our system provides clear traceability through explicit tool selection and parameter extraction, allowing users to observe which tool was selected, what parameters were extracted, and why validation succeeded or failed. \textbf{(3) Modular Design:} the system employs a modular architecture with separate functions for different capabilities, loading only relevant tools at runtime to improve efficiency. \textbf{(4) Customization:} users can easily adapt the system by defining custom tools for specific applications, replacing the underlying control model, or extending the interface to support different modalities such as simulation environments.

\subsection{Complete Adaptation Workflow}

The complete workflow is formalized in \Cref{algorithm:llm_trajectory_adaptation}, integrating tool-based architecture, tool selection and validation, and KMP adaptation tools. We denote the LLM as $\mathcal{L}$, which maps natural language instructions to tool selections and parameters.
\begin{algorithm}
    \caption{\emph{LLM-based trajectory adaptation with natural language feedback}}
    \begin{algorithmic}[1]
        \State{\textbf{Input}: $\kmp = \{\kmpInput, \kmpMean, \kmpCovariance\}_{\kmpIndex=1}^\amountOfKMP$ from $\mathcal{D}$, LLM $\mathcal{L}$, tool set $\mathcal{T}$, environment $\mathcal{E}$}

        \While{user interaction continues}
            \State $\outputVariable(\inputVariable) \gets \text{Predict}(\kmp)$ via Eqs.~\eqref{eq:kmp_mean}--\eqref{eq:kmp_cov} \Comment{Execute trajectory}

            \State Receive $L_{\text{instruct}}$ from user \Comment{Natural language instruction}

            \State $T^* \gets \mathcal{L}_{\text{select}}(L_{\text{instruct}}, \mathcal{E}, \mathcal{T})$ \Comment{Select tool from $\mathcal{T}$}

            \State $\theta^* \gets \mathcal{L}_{\text{param}}(L_{\text{instruct}}, \mathcal{E}, T^*)$ \Comment{Extract parameters}

            \If{$\text{Validate}(\theta^*, \kmp)$ fails} \Comment{Type, range, workspace, physics checks}
                \State Provide feedback and \textbf{continue}
            \EndIf

            \State $\kmp \gets T^*(\kmp, \theta^*)$ \Comment{Apply tool modification:}

            \LeftComment{Speed: ${\timeInterval'}_{\trajectoryIndex} = {\timeInterval}_{\trajectoryIndex} \cdot f(\gamma)$ for $\inputVariable \in [\inputVariable_{\text{start}}, \inputVariable_{\text{end}}]$}

            \LeftComment{Via-point: $\kmp \gets \kmp \cup \{\viapointInput, \viapointMean, \viapointCovariance\}$}

            \LeftComment{Repulsion: Add via-points $\{\inputVariable_{\trajectoryIndex}, \bm{x}'(\inputVariable_{\trajectoryIndex})\}$ where $d(\mathbb{E}[\bm{\xi}(\inputVariable_{\trajectoryIndex})]) < \delta_{\text{safe}}$}

        \EndWhile

        \State{\textbf{Output}: Adapted $\kmp$ for robot execution}

    \end{algorithmic}
    \label{algorithm:llm_trajectory_adaptation}
\end{algorithm}
The algorithm shows how the LLM selects and parameterizes well-defined tools that modify the KMP's internal trajectory representation.

	\section{Evaluation}
	\label{sec:evaluation}
	\subsection{Evaluation on real robot}\label{subsec:eval_on_robot}
\begin{figure}
    \centering
    \includegraphics[width=0.61\columnwidth]{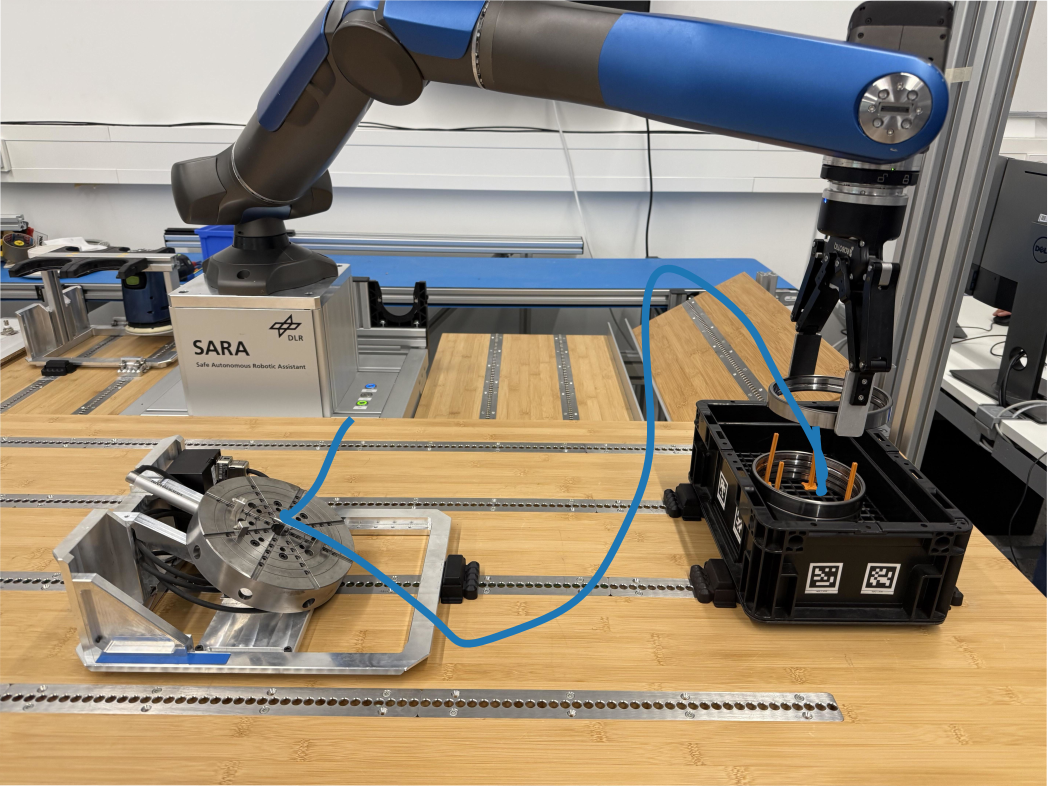}
    \includegraphics[width=1.0\columnwidth]{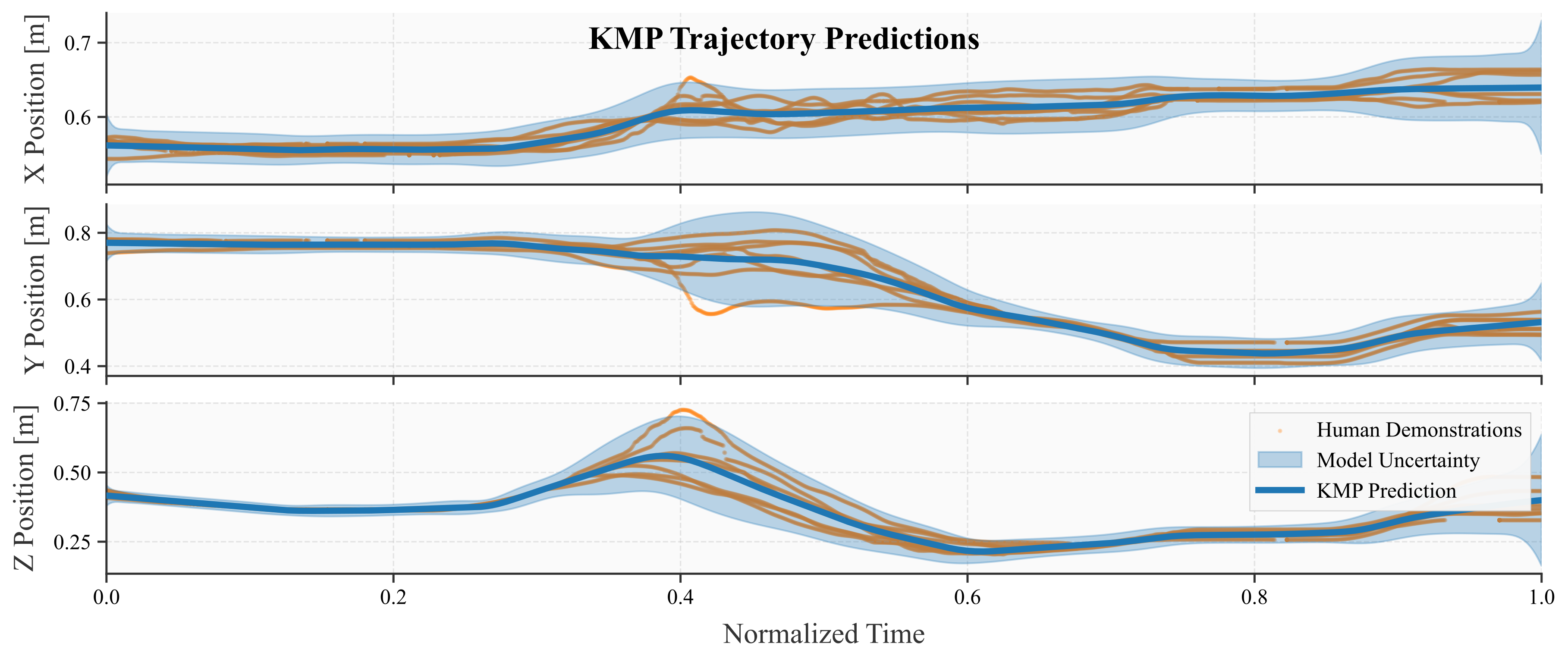}
    \caption{Demonstration and prediction analysis for the pick-and-insert task. (Top) Trajectory in robot frame. (Bottom) Demonstrations and KMP model predictions with uncertainty visualization.}
    \label{fig:demonstration_analysis}
\end{figure}

We evaluate our approach on a torque-controlled 7-DoF robot in an industrial scenario where an inner bearing ring must be transferred from its box to a measurement device on a workbench. We use Qwen2.5-VL-72B-Instruct as the LLM backend.
We provide $\amountOfDatapoints=6$ demonstrations for training the KMP model (\Cref{fig:demonstration_analysis}).
We use a time-driven representation with $\inputVariable_{\trajectoryIndex,\datapointIndex}=t_{\trajectoryIndex,\datapointIndex}/T_{\datapointIndex}$, where $t$ is a time step, and learn the end-effector position $\outputVariable_{\trajectoryIndex,\datapointIndex}=\bm{x}_{\trajectoryIndex,\datapointIndex}$.
We map all the inputs to the interval $[0,1]$ through division by the duration of each demonstration $T_{\datapointIndex}$ to easily re-scale skill duration.
The KMP is initialized from a GMM with 12 components and $\trajectoryLength=500$ inputs.
We chose a Mat\'ern kernel ($\nu=5/2$) with length scale $l=0.1$, noise variance 1.0~\cite{Rasmussen06}, and KMP hyperparameters $\lambda_1=0.1$, $\lambda_2=1$, $\alpha=1$, chosen empirically.
The experiments are also shown in the accompanying video.

\textbf{Evaluation Metrics:} We assess our system using \textit{Command Success Rate (CSR)}, \textit{Interpretation Success Rate (ISR)}, and \textit{Task Completion Rate (TCR)}---representing the percentage of executions without failures, correctly interpreted commands, and successfully completed tasks, respectively---alongside \textit{Response Time}, measuring average overall adaptation time.



%
\subsection{Experiment 1: Speed adaptation via natural language}
\label{subsec:speed_adaptation}

\begin{figure}
    \centering
    \includegraphics[width=1.0\columnwidth]{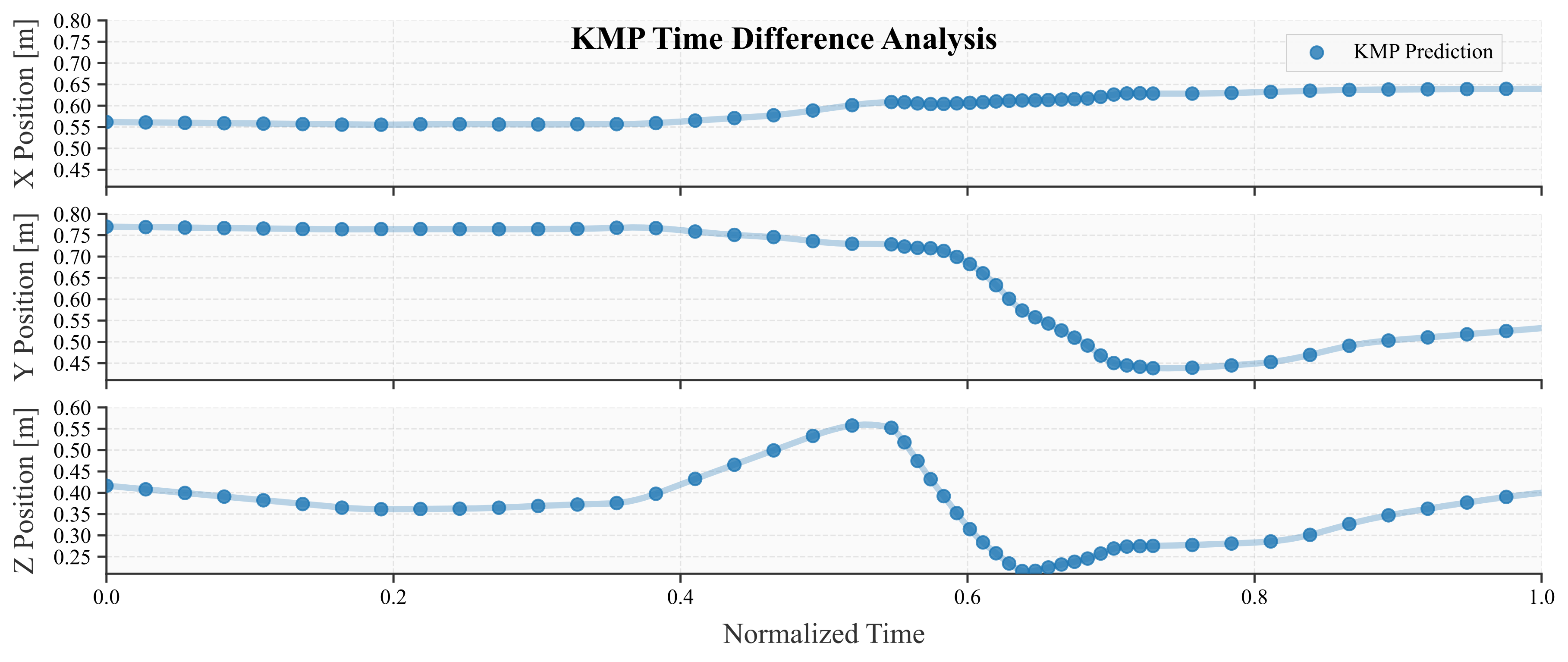}
    \caption{Speed adaptation results showing temporal trajectory modification through natural language commands. The plot shows the adapted trajectories following command ``slow down between box and station,'' demonstrating precise temporal control while preserving spatial trajectory characteristics.}
    \label{fig:speed_adaptation}
\end{figure}

This experiment evaluates the system's ability to modify execution speed through natural language commands.
The robot initially performs the pick-and-insert task at a baseline speed learned from demonstrations.
The user provides the command ``slow down between box and station''.
The environment observations $\mathcal{E}$ include object positions and semantic labels, provided manually. Dense-pose estimation~\cite{Sundermeyer2018} could alternatively be used for automatic perception.
Using the trajectory segment determination method described in \Cref{sec:approach} with proximity threshold $d_{\text{prox}} = 2$~cm, the system identifies the temporal boundaries for adaptation.

The system applies the speed modulation tool (see \Cref{sec:approach}) to adjust the time intervals ${\timeInterval}_{\trajectoryIndex}$ between time steps $t=0.55$ and $t=0.72$, corresponding to the phase after picking up the ring and before reaching the measurement station.
As seen in \Cref{fig:speed_adaptation}, the adapted trajectory successfully slows down in the specified region while preserving the spatial characteristics, demonstrating robust natural language understanding for temporal skill adaptation.
Our approach achieved perfect performance on this task with 100\% CSR, ISR, and TCR across all test variations (\Cref{tab:task_comparison_ovita}).

\subsection{Experiment 2: Trajectory correction through language feedback}
\label{subsec:trajectory_correction}

\begin{figure}
	\centering
	\includegraphics[width=0.61\columnwidth]{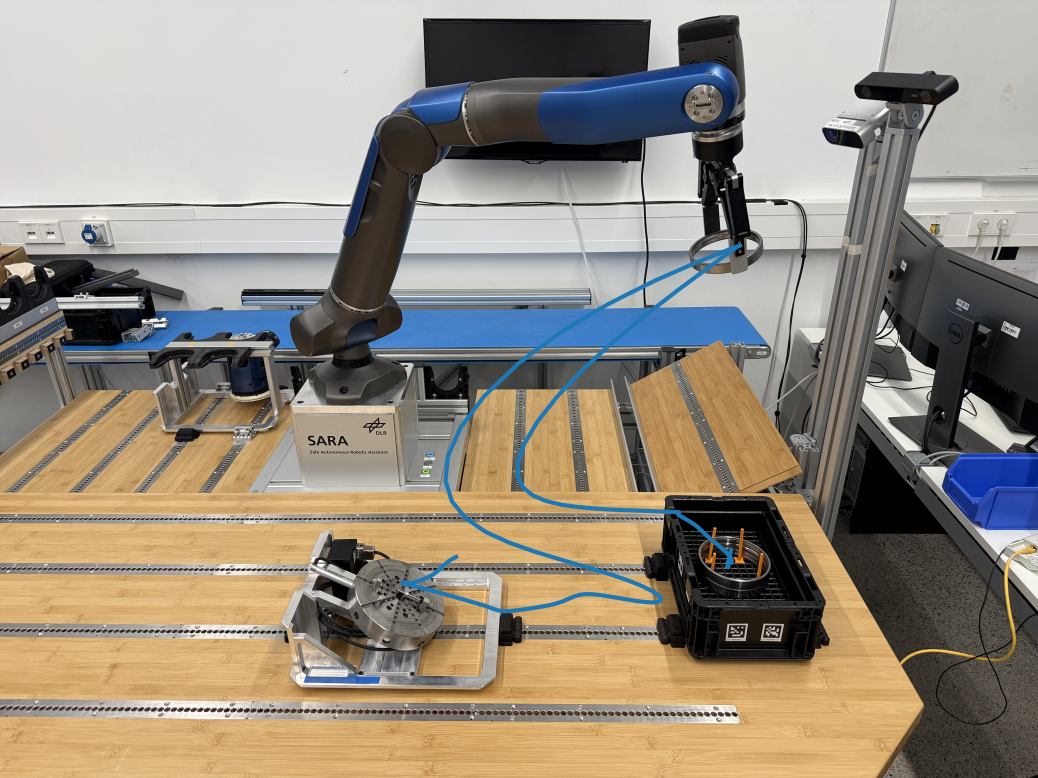}
	\includegraphics[width=1.0\columnwidth]{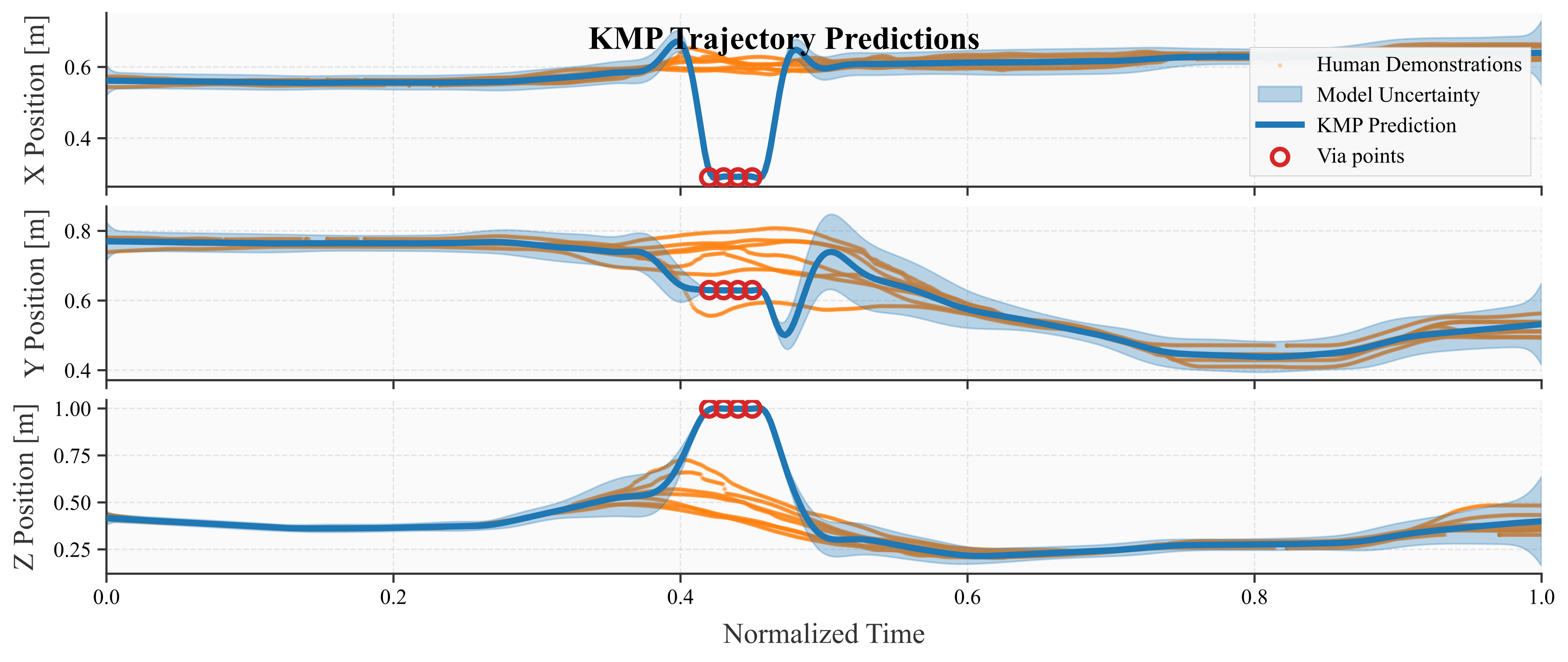}
	\caption{Trajectory adaptation through natural language command showing (top) experimental setup with new object (camera) placement and (bottom) resulting KMP trajectory modification with via-point insertion to reach the new object while maintaining task completion.}
	\label{fig:trajectory_adaptation}
\end{figure}

After the baseline demonstrations, we introduce a new object to the workspace: a camera positioned to the left of the robot (visible in \Cref{fig:trajectory_adaptation}, top).
The user provides the command ``Check the ring with the camera on the left''.
The LLM identifies the camera object from $\mathcal{E}$, extracting its position $\bm{p}_{\text{camera}}$ and semantic label.
Using the trajectory segment determination method described in \Cref{sec:approach}, the system determines when the ring is picked up.
The via-point insertion tool (see \Cref{sec:approach}) then places a via-point at the camera location, with temporal placement between this identified pickup time and the insertion at the measurement station.
\Cref{fig:trajectory_adaptation} (bottom) shows the adapted trajectory with the via-point at the camera location.
The robot successfully executes the modified trajectory, demonstrating that the system can incorporate new task requirements while maintaining the learned skill structure.
Our approach achieved 100\% CSR and TCR, with 80\% ISR (\Cref{tab:task_comparison_ovita}). The reduced ISR occurred when the LLM additionally applied speed modulation alongside via-point insertion, which, while functional, was not explicitly requested.

\subsection{Experiment 3: Avoiding novel obstacles}
\label{subsec:obstacle_avoidance}

\begin{figure}
	\centering
	\includegraphics[width=0.61\columnwidth]{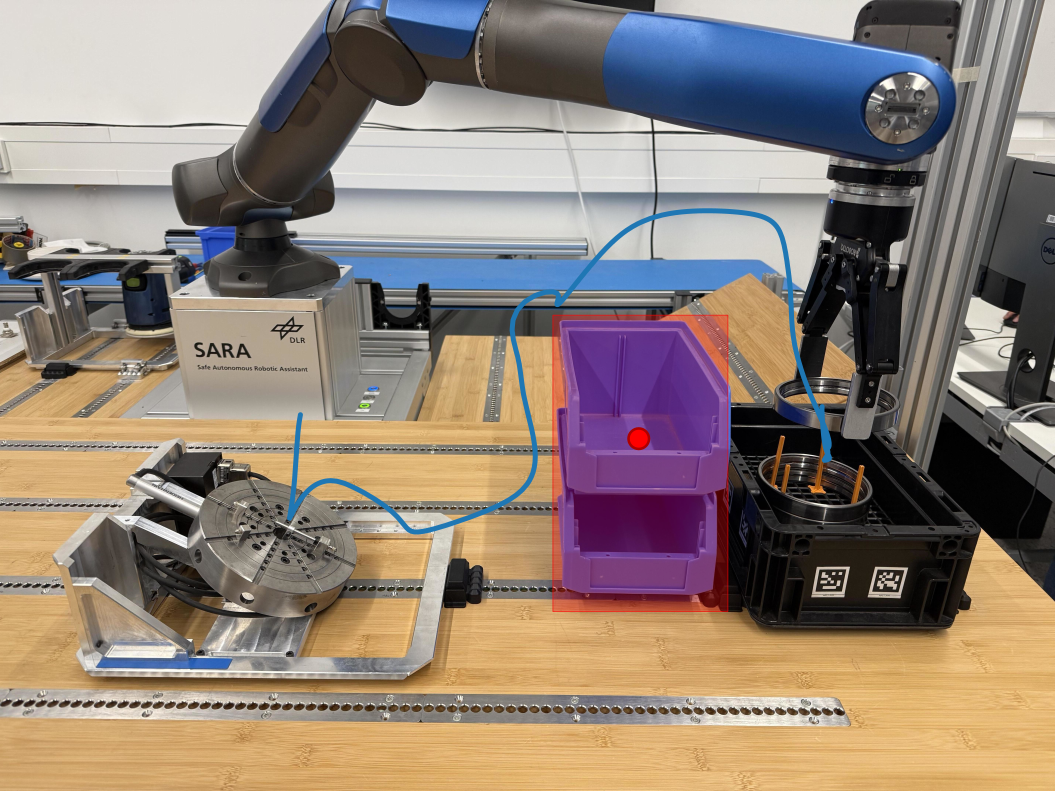}
	\includegraphics[width=1.0\columnwidth]{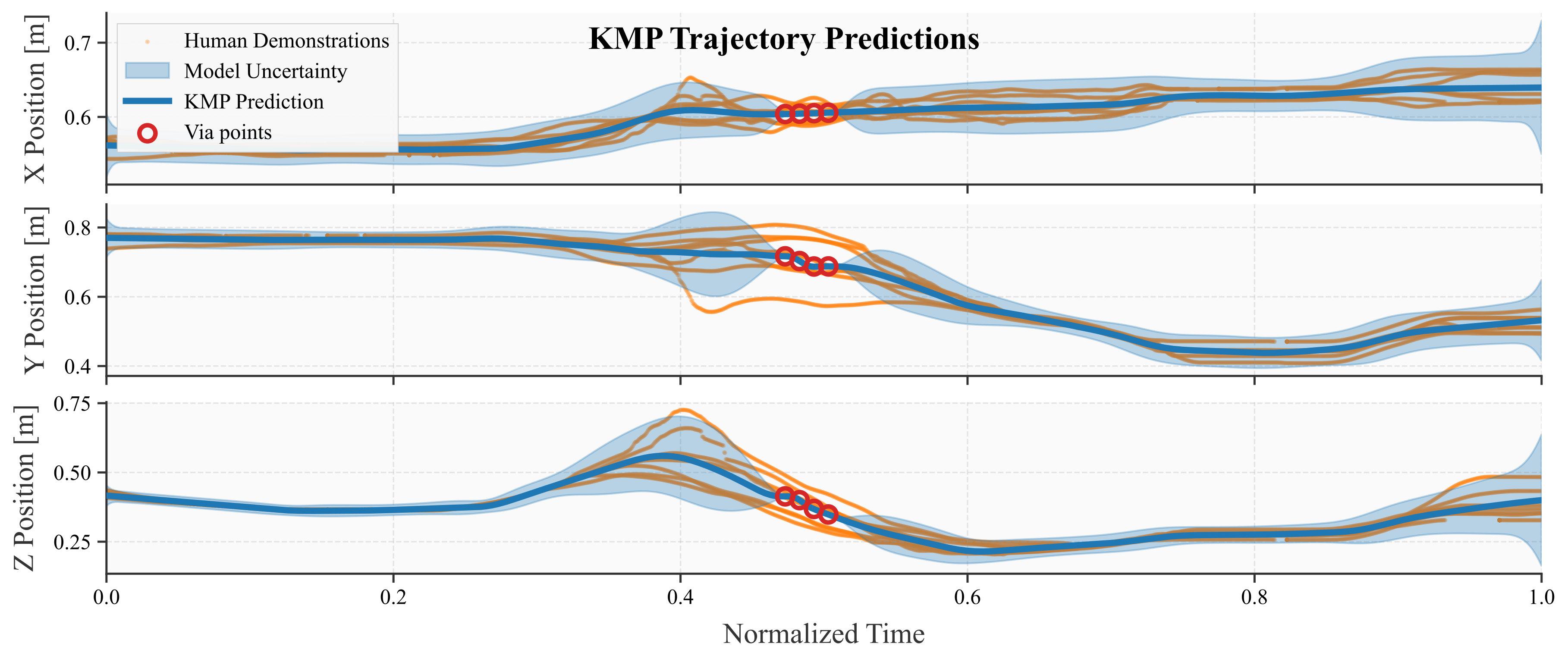}
	\caption{Obstacle avoidance through natural language command showing (top) experimental workspace with introduced obstacle and (bottom) KMP trajectory modification using repulsion points. The system generates collision-free paths while maintaining the overall task structure and successfully completing the pick-and-insert operation.}
	\label{fig:obstacle_avoidance}
\end{figure}

This experiment evaluates the system's ability to adapt trajectories to avoid obstacles introduced to the workspace after the initial demonstrations.
A blue industrial box is placed next to the box where the robot picks up the bearing ring, creating a collision hazard with the original learned trajectory (visible in \Cref{fig:obstacle_avoidance}, top).
The user provides the command ``Please avoid the blue box''.
The LLM identifies the obstacle from $\mathcal{E}$, extracting its position $\bm{p}_{\text{obstacle}}$, dimensions $\bm{d}_{\text{obstacle}}$, and semantic label (provided manually).
The repulsion point generation tool constructs a signed distance field (SDF) representing the obstacle volume from its dimensions (see \Cref{sec:approach} for SDF formulation). The SDF is modeled using the box dimensions ($15 \times 22.5 \times 23.5$~cm) with safety margin $\delta_{\text{safe}} = 2$~cm. The tool evaluates the predicted trajectory for potential collisions and inserts via-points along the affected segment where corrections are needed, generating a collision-free path that passes over the blue box.
\Cref{fig:obstacle_avoidance} (bottom) shows the adapted trajectory with the generated repulsion points.
The robot successfully avoids the obstacle while completing the task, achieving 100\% CSR, ISR, and TCR (\Cref{tab:task_comparison_ovita}).

\subsection{Comparison with OVITA}
\label{subsec:comparison_ovita}

We compare our approach with OVITA~\cite{Maurya2025}, a recent open-vocabulary method for trajectory adaptation. While both systems enable language-driven trajectory modifications, they differ fundamentally in their underlying representations and adaptation mechanisms.
We evaluate OVITA in two configurations: (1) the original implementation using cloud-based LLMs (we use Gemini), and (2) a modified version using the same local LLM deployed in our approach (Qwen2.5-VL-72B-Instruct) to ensure fair comparison under identical language understanding capabilities. This dual evaluation isolates the contributions of the underlying trajectory representation and adaptation mechanism from differences in LLM performance.
The results are shown in \Cref{tab:comparison_ovita,tab:task_comparison_ovita}.
OVITA with cloud-based LLMs achieved performance consistent with its original publication, demonstrating effective code generation across all scenarios. However, transitioning to a local LLM revealed significant performance degradation, particularly in speed adaptation and trajectory correction. Low ISR values indicate that generated code successfully modified trajectories but failed to match user intent. Degraded TCR values reflect substantial deviations from task-critical waypoints, preventing successful task completion. In contrast, our tool-based approach maintains consistent performance with the same local LLM, demonstrating the robustness of structured tool interfaces over code generation.

\begin{table}[h]
\caption{Comparison with OVITA baseline}
\centering
\begin{tabular}{l c}
\toprule
\textbf{Method} & \textbf{Time (s)} \\ 
\midrule
OVITA - LLM: Gemini (cloud) & 72.1 \\
OVITA - LLM: Qwen2.5-VL-72B-Instruct (local) & 27.1 \\
\textbf{IROSA} - LLM: Qwen2.5-VL-72B-Instruct (local) & \textbf{15.4} \\
\bottomrule
\end{tabular}%
\label{tab:comparison_ovita}
\end{table}

\begin{table}[h]
\caption{Quantitative comparison across experiments (10 prompts $\times$ 5 runs each). Prompts collected from public demonstration visitors and colleagues. Full prompt set available in our repository.}
\centering
\resizebox{\columnwidth}{!}{%
\begin{tabular}{l c c c c c c c c c}
\toprule
\textbf{Experiment} & \multicolumn{3}{c}{\textbf{OVITA (cloud)}} & \multicolumn{3}{c}{\textbf{OVITA (local)}} & \multicolumn{3}{c}{\textbf{IROSA (local)}} \\
\cmidrule(lr){2-4} \cmidrule(lr){5-7} \cmidrule(lr){8-10}
 & \textbf{CSR} & \textbf{ISR} & \textbf{TCR} & \textbf{CSR} & \textbf{ISR} & \textbf{TCR} & \textbf{CSR} & \textbf{ISR} & \textbf{TCR} \\
\midrule
Speed Adapt. & \textbf{100.0} & \textbf{100.0} & \textbf{100.0} & 70.0 & 10.0 & 30.0 & \textbf{100.0} & \textbf{100.0} & \textbf{100.0} \\
Traj. Correct. & 90.0 & 40.0 & 30.0 & 40.0 & 10.0 & 0.0 & \textbf{100.0} & \textbf{80.0} & \textbf{100.0} \\
Obst. Avoid. & 80.0 & 40.0 & 30.0 & 60.0 & 60.0 & 0.0 & \textbf{100.0} & \textbf{100.0} & \textbf{100.0} \\
\bottomrule
\end{tabular}%
}
\label{tab:task_comparison_ovita}
\end{table}

	\section{Discussion}
	\label{sec:discussion}
	\subsection{Analysis of Results}

Our experiments demonstrate that users can effectively modify robot skills through natural language without specialized training. The system interprets diverse commands, from relative (``faster'') to absolute (``20\% faster''), and translates them into appropriate KMP modifications while preserving skill structure.
Compared to OVITA, our approach achieves superior performance across all metrics (CSR, ISR, TCR) while reducing response time by 43\% when using the same local LLM, demonstrating that structured tool interfaces outperform code generation for skill adaptation.

\subsection{Scalability Analysis and Performance Considerations}
We tested our approach with different local LLM models (\textit{Qwen2.5-Omni-7B}, \textit{Qwen2.5-VL-72B-Instruct}, and \textit{Pixtral-12B-2409}) and experienced the following points:

\textbf{Tool set scaling} faces context length constraints when exceeding 15 tools with smaller LLMs (\eg 8K context models), where extensive tool descriptions can exceed prompt limits and lead to processing errors. Our modular approach addresses this by dynamically loading only task-relevant tools, maintaining performance while staying within context boundaries. Larger models can handle more tools simultaneously, indicating that scaling limitations are primarily determined by the LLM's context capacity rather than our architectural design.
\textbf{Language model scaling} benefits from larger models' improved natural language understanding capabilities. However, the tool-based interface provides consistent behavior across different model sizes, as tool selection relies on semantic matching rather than model-specific training. This design enables system deployment with various LLM backends without architecture modifications.

\subsection{Flexibility-Safety Trade-off}
Our tool-based approach trades flexibility for safety and predictability. OVITA's code generation offers greater flexibility for open-ended adaptations (\eg ``execute a spiral'') that our predefined tools cannot express. Our approach targets industrial scenarios where certification and operator trust are paramount. New tools can expand capabilities while preserving safety guarantees.
For instructions requiring multiple modifications (\eg ``slow down near the station and avoid the blue box''), the LLM can select multiple tools from a single command; however, the system is limited to compositions of available tools.
We observed cases where the LLM applied additional modifications not explicitly requested. While these remained within safe bounds, they represent unintended behavior. Stricter prompt engineering and post-processing filters that flag multi-tool responses for user review can mitigate this.
Parameter validation within tools and retry mechanisms handle edge cases where the LLM produces invalid outputs.

\subsection{Limitations and Future Work}
While our approach demonstrates promising results, several areas warrant further investigation. The current system relies on pre-defined tool functions, which constrains adaptability to user intents not captured by the existing tool set. Future work could explore automatic tool generation or more flexible function composition mechanisms to broaden system capabilities.
The framework currently focuses on adapting a single given skill through natural language commands but does not address selection and switching between different skills or primitives. Future research should investigate natural language-based skill selection and composition mechanisms to enable broader robot behavior adaptation.
The interpretability of natural language feedback for end-users was not experimentally evaluated; assessing whether users can verify their intent from feedback alone remains future work.
The modular architecture enables choosing whether physical interaction~\cite{Knauer2025} or natural language is used to achieve skill adaptations. Future research should investigate which modality is better suited for different scenarios and user groups, and how to effectively combine both approaches.

	\section{Conclusion}
	\label{sec:conclusion}
	We presented a novel framework for natural language-based robot skill adaptation using a tool-based architecture that maintains strict separation between language understanding and robot control.
The framework leverages local, pre-trained LLMs without additional training, interfacing with Kernelized Movement Primitives through validated tool functions to enable speed adjustments, trajectory corrections, and obstacle avoidance through everyday language.
Experimental validation on a 7-DoF robot performing industrial manipulation demonstrates effective interpretation of diverse commands while preserving skill structure and ensuring predictable execution.



	\bibliographystyle{IEEEtran}
	\bibliography{bibliography.bib}
	
\end{document}